\documentclass[11pt,a4paper]{article}
\usepackage[hyperref]{emnlp2018}
\usepackage{times}
\usepackage{latexsym}

\usepackage{amsfonts}
\usepackage{amsmath}

\usepackage[pdftex]{graphicx}
\usepackage{CJKutf8}

\usepackage{subcaption}

\usepackage{multirow}
\usepackage{url}

\aclfinalcopy 
\def\aclpaperid{137} 

\title{The University of Cambridge’s Machine Translation Systems for WMT18}

\author{Felix Stahlberg$^\dag$ \and Adri\`a de Gispert$^{\ddagger\dag}$ \and Bill Byrne$^{\ddagger\dag}$\\
 $^\dag$Department of Engineering, University of Cambridge, UK  \\
 $^\ddagger$SDL Research, Cambridge, UK \\
  {\tt \{fs439,ad465,wjb31\}@cam.ac.uk} \\
  }

\date{}

\begin{document}
\maketitle
\begin{abstract}
The University of Cambridge submission to the WMT18 news translation task focuses on the combination of diverse models of translation. We compare recurrent, convolutional, and self-attention-based neural models on German-English, English-German, and Chinese-English. Our final system combines all neural models together with a phrase-based SMT system in an MBR-based scheme. We report small but consistent gains on top of strong Transformer ensembles.
\end{abstract}

\section{Introduction}

Encoder-decoder networks~\citep{early-raam1,early-raam2,early-raam3,encdec} are the current prevailing architecture for neural machine translation (NMT). Various architectures have been used in the general framework of encoder and decoder networks such as recursive auto-encoders~\citep{early-raam1,socher-recursive-autoencoders,li-recursive-autoencoders}, (attentional) recurrent models~\citep{sutskever,bahdanau,dot-product-attention,gnmt,rnmt}, convolutional models~\citep{encdec,slicenet,convs2s}, and, most recently, purely (self-)attention-based models~\citep{transformer,weighted-transformer,relative-transformer}. In the spirit of \citet{rnmt} we devoted our WMT18 submission to exploring the three most commonly used architectures: recurrent, convolutional, and self-attention-based models like the Transformer~\citep{transformer}. Our experiments suggest that self-attention is the superior architecture on the tested language pairs, but it can still benefit from model combination with the other two. We show that using large batch sizes is crucial to Transformer training, and that the delayed SGD updates technique~\citep{danielle-syntax} is useful to increase the batch size on limited GPU hardware. Furthermore, we also report gains from MBR-based combination with a phrase-based SMT system. We found this particularly striking as the SMT baselines are often more than 10 BLEU points below our strongest neural models. Our final submission ranks second in terms of BLEU score in the WMT18 evaluation campaign on English-German and German-English, and outperforms all other systems on a variety of linguistic phenomena on German-English~\citep{wmt18-deen-ana}.

\section{System Combination}
\label{sec:sys-combination}

\citet{mbr-nmt} combined SMT and NMT in a hybrid system with a minimum Bayes-risk (MBR) formulation which has been proven useful even for practical industry-level MT~\citep{hybrid-mbr-sdl}. Our system combination scheme is a generalization of this approach to more than two systems. Suppose we want to combine $q$ models $\mathcal{M}_1,\dots,\mathcal{M}_q$. We first divide the models into two groups by selecting a $p$ with $1\leq p\leq q$. We refer to scores from the first group $\mathcal{M}_1,\dots,\mathcal{M}_p$ as {\em full posterior} scores and from the second group $\mathcal{M}_{p+1},\dots,\mathcal{M}_q$ as {\em MBR-based} scores. Full posterior models contribute to the combined score with their complete posterior of the full translation. In contrast, models in the second group only provide the evidence space for estimating the probability of $n$-grams occurring in the translation. Full-posterior models need to assign scores via the standard left-to-right factorization of neural sequence models:

\begin{equation}
\log P(y_1^T|\mathbf{x},\mathcal{M}_i)=\sum_{t=1}^T \log P(y_t|y_1^{t-1},\mathbf{x},\mathcal{M}_i)
\end{equation}
for a target sentence $\mathbf{y}=y_1^T$ of length $T$ given a source sentence $\mathbf{x}$ for all $i\leq p$. For example, all left-to-right neural models in this work can be used as full posterior models, but the right-to-left models (Sec.~\ref{sec:right-to-left}) and SMT cannot. We combine full-posterior scores log-linearly, and bias the combined score $S(\mathbf{y}|\mathbf{x})$ towards low-risk hypotheses with respect to the MBR-based group as suggested by \citet[Eq.~4]{mbr-nmt}:\footnote{Eq.~\ref{eq:combined-score} differs from Eq.~4 of \citet{mbr-nmt} in that we do not use a word penalty $\Theta_0$ here, and we do not tune weights for different order $n$-grams separately ($\Theta_1,\dots\Theta_4$). Both did not improve translation quality in our setting.}

\begin{equation}
\begin{aligned}
S(\mathbf{y}|\mathbf{x})=\sum_{t=1}^T \Big( &\underbrace{\sum_{i=1}^p \lambda_i\log P(y_t|y_1^{t-1},\mathbf{x},\mathcal{M}_i)}_{\text{Full posterior}}+ \\
&\underbrace{\sum_{j=p+1}^q \lambda_j \sum_{n=1}^4 P(y_{t-n}^t|\mathbf{x},\mathcal{M}_j)}_{\text{MBR-based $n$-gram scores}}\Big)
\label{eq:combined-score}
\end{aligned}
\end{equation}
where $\lambda_1,\dots,\lambda_q$ are interpolation weights. Eq.~\ref{eq:combined-score} also describes how to use beam search in this framework as hypotheses can be built up from left to right due to the outer sum over time steps. The MBR-based models contribute via the probability $P(y_{t-n}^t|\mathbf{x},\mathcal{M}_j)$ of an $n$-gram $y_{t-n}^t$ given the source sentence $\mathbf{x}$. Posteriors in this form are commonly used for MBR decoding in SMT~\citep{mbr-shankar,mbr-tromble}, and can be extracted efficiently from translation lattices using counting transducers~\citep{lmbr-hifst}. For our neural models we run beam search with beam size 15 and compute posteriors over the 15-best list. We smooth all $n$-gram posteriors as suggested by~\citet{mbr-nmt}.

Note that our generalization to more than two systems can still be seen as instance of the original scheme from~\citet{mbr-nmt} by viewing the first group $\mathcal{M}_1,\dots,\mathcal{M}_p$ as ensemble and the evidence space from the second group $\mathcal{M}_{p+1},\dots,\mathcal{M}_q$ as mixture model.

The performance of our system combinations depends on the correct calibration of the interpolation weights $\lambda_1,\dots,\lambda_q$. We first tried to use $n$-best or lattice MERT~\citep{nbest-mert,lmert} to find interpolation weights, but these techniques were not effective in our setting, possibly due to the lack of diversity and depth in $n$-best lists from standard beam search. Therefore, we tune on the first best translation using Powell's method~\citep{powell} with a line search algorithm similar to golden-section search~\citep{golden-section-search}.

\section{Right-to-left Translation Models}
\label{sec:right-to-left}

Standard NMT models generate the translation from left to right on the target side. Recent work has shown that incorporating models which generate the target sentence in reverse order (i.e. from right to left) can improve translation quality~\citep{right-to-left-agreement,right-to-left-draft,uedin-wmt17,microsoft-parity}. Right-to-left models are often used to rescore $n$-best lists from left-to-right models. However, we could not find improvements from rescoring in our setting. Instead, we extract $n$-gram posteriors from the R2L model, reverse them, and use them for system combination as described in Sec.~\ref{sec:sys-combination}. 

\section{Experimental Setup}

\subsection{Data Selection}

We ran language detection~\citep{langdetect} and gentle length filtering based on the number of characters and words in a sentence on all available monolingual and parallel data in English, German, and Chinese. Due to the high level of noise in the ParaCrawl corpus and its large size compared to the rest of the English-German data we additionally filtered ParaCrawl more aggressively with the following rules:

\begin{itemize}
\item No words contain more than 40 characters.
\item Sentences must not contain HTML tags.
\item The minimum sentence length is 4 words.
\item The character ratio between source and target must not exceed 1:3 or 3:1.
\item Source and target sentences must be equal after stripping out non-numerical characters.
\item Sentences must end with punctuation marks.
\end{itemize}

This additional filtering reduced the size of ParaCrawl from originally 36M sentences to 19M sentences after language detection, and to 11M sentences after applying the more aggressive rules.

For backtranslation~\citep{backtranslation} we selected 20M sentences from News Crawl 2017. 
We used a single Transformer~\citep{transformer} model in Tensor2Tensor's~\citep{t2t} \texttt{transformer\_base} configuration for generating the synthetic source sentences. We over-sampled~\citep{uedin-wmt17} WMT data by factor 2 except the ParaCrawl data and the UN data on Chinese-English to roughly match the size of the synthetic data. Tabs.~\ref{tab:train-data-size-ende} and~\ref{tab:train-data-size-zhen} summarize the sizes of our final training corpora.

\begin{table}[t!]
\begin{center}
\small
\begin{tabular}{|l|r|r|}
\hline
\textbf{Corpus} & \textbf{Over-sampling} & \textbf{\#Sentences} \\
\hline
Common Crawl & 2x & 4.43M \\
Europarl v7 & 2x & 3.76M \\
News Commentary v13 & 2x & 0.57M \\
Rapid 2016 & 2x & 2.27M \\
ParaCrawl & 1x & 11.16M \\
Synthetic (news-2017) & 1x & 20.00M \\
\hline
\textbf{Total} & & \textbf{42.19M} \\
\hline
\end{tabular}
\end{center}
\caption{\label{tab:train-data-size-ende} Training data sizes for English-German and German-English after filtering.}
\end{table}

\begin{table}[t!]
\begin{center}
\small
\begin{tabular}{|l|r|r|}
\hline
\textbf{Corpus} & \textbf{Over-sampling} & \textbf{\#Sentences} \\
\hline
CWMT - CASIA2015 & 2x & 2.08M \\
CWMT - CASICT2015 & 2x & 3.95M \\
CWMT - Datum2017 & 2x & 1.93M \\
CWMT - NEU2017 & 2x &  3.95M \\
News Commentary v13 & 2x & 0.49M \\
UN v1.0 & 1x & 14.25M \\
Synthetic (news-2017) & 1x & 20.00M \\
\hline
\textbf{Total} & & \textbf{46.66M} \\
\hline
\end{tabular}
\end{center}
\caption{\label{tab:train-data-size-zhen} Training data sizes for Chinese-English after filtering.}
\end{table}

\subsection{Preprocessing}

We preprocess our English and German data with Moses tokenization, punctuation normalization, and truecasing. On Chinese we first used the WMT \texttt{tokenizeChinese.py}\footnote{\url{http://www.statmt.org/wmt17/tokenizeChinese.py}} script and separated segments of Chinese and Latin text from each other. Then, we removed whitespace between Chinese characters and tokenized Chinese segments with Jieba\footnote{\url{https://github.com/fxsjy/jieba}} and the rest with \texttt{mteval-v13a.pl}. For our neural models we apply byte-pair encoding~\citep[BPE]{bpe} with 32K merge operations. We use joint BPE vocabularies on English-German and German-English and separate source/target encodings on Chinese-English.

\subsection{Model Hyper-Parameters}

We use 1024-dimensional embedding and output projection layers in all architectures. The embeddings are shared between encoder and decoder on English-German and German-English, but not on Chinese-English.

\begin{table}[t!]
\begin{center}
\small
\begin{tabular}{|l|r|r|r|r|r|}
\hline
\textbf{Architecture} & \textbf{en-de, de-en} &\multicolumn{1}{c|}{\textbf{zh-en}}  \\
\hline
LSTM & 114.2M & 192.7M \\
SliceNet & 27.5M & 86.4M \\
Transformer & 212.8M & 291.4M \\
Relative Transformer & 213.8M & 292.5M\\
\hline
\end{tabular}
\end{center}
\caption{\label{tab:params} Number of model parameters.}
\end{table}

\paragraph{LSTM}

For our recurrent models we adapted the TensorFlow seq2seq tutorial code base~\citep{tfnmt} for use inside the Tensor2Tensor library~\citep{t2t}.\footnote{\url{https://github.com/fstahlberg/tensor2tensor-usr}} We roughly followed the UEdin WMT17 submission~\citep{uedin-wmt17} and stacked four 1024-dimensional LSTM layers with layer normalization~\citep{layernorm} and residual connections in both the decoder and bidirectional encoder. We equipped the decoder network with Bahdanau-style~\citep{bahdanau} attention (\texttt{normed\_bahdanau}).

\begin{table}[t!]
\begin{center}
\small
\begin{tabular}{|r|r|r|r|r|}
\hline
\textbf{\#Physical} & \multicolumn{1}{c|}{\textbf{Delay}} & \multicolumn{1}{c|}{\textbf{\#Effective}} & \textbf{Effective} &  \\
\multicolumn{1}{|c|}{\textbf{GPUs}} & \textbf{factor} & \multicolumn{1}{c|}{\textbf{GPUs}} & \textbf{batch size} &  \textbf{BLEU} \\
\multicolumn{1}{|c|}{($g$)} & \multicolumn{1}{c|}{($d$)} & \multicolumn{1}{c|}{(g'=gd)} & \multicolumn{1}{c|}{(b'=bg')} &  \\
\hline
1 & 1 & 1 & 2,048 & 28.2 \\
4 & 1 & 4 & 8,192 & 29.5 \\
4 & 4 & 16 & 32,768 & 30.3 \\
4 & 16 & 64 & 131,072 & 29.8 \\
\hline
\end{tabular}
\end{center}
\caption{\label{tab:batch-size} Impact of the effective batch size on Transformer training on en-de news-test2017 after 3,276M training tokens, beam size 4.}
\end{table}

\paragraph{SliceNet}

The convolutional model of \citet{slicenet} called SliceNet is implemented in Tensor2Tensor. We use the standard configuration \texttt{slicenet\_1} of four hidden layers with layer normalization.

\paragraph{Transformer} We compare two Transformer variants available in Tensor2Tensor: the original Transformer~\citep{transformer} (\texttt{transformer\_big} setup) and the Transformer of \citet{relative-transformer} with relative positional embeddings (\texttt{transformer\_relative\_big} setup). Both use 16-head dot-product attention and six 1024-dimensional encoder and decoder layers.

The number of training parameters of our neural models is summarized in Tab.~\ref{tab:params}.

\begin{table*}[t!]
\begin{center}
\small
\begin{tabular}{|l|r|r|r|r|r|}
\hline
\textbf{Architecture} & \textbf{\#Effective GPUs} & \textbf{Batch size} & \textbf{\#SGD updates} & \textbf{\#Training tokens}  \\
\hline
LSTM & 8 & 4,096 & 45K & 1,475M \\
SliceNet & 4 & 2,048 & 800K & 6,554M \\
R2L Transformer & 16 & 2,048 & 200K & 6,554M \\
Transformer & 16 & 2,048 & 250K & 8,192M \\
Relative Transformer & 16 & 2,048 & 250K & 8,192M \\
\hline
\end{tabular}
\end{center}
\caption{\label{tab:training} Training setups for our neural models on all language pairs.}
\end{table*}

\subsection{Training}

We train vanilla phrase-based SMT systems\footnote{Excluding the UN corpus and the backtranslated data.} and extract 1000-best lists of unique translations candidates, from which $n$-gram posteriors are calculated.

All neural models were trained with the Adam optimizer~\citep{adam}, dropout~\citep{dropout}, and label smoothing~\citep{label-smoothing} using the Tensor2Tensor~\citep{t2t} library. We decode with the average of the last 40 checkpoints~\citep{marian}.

We make extensive use of the delayed SGD updates technique we already applied successfully to syntax-based NMT~\citep{danielle-syntax}. Delaying SGD updates allows to arbitrarily choose the effective batch size even on limited GPU hardware. Large batch training has received some attention in recent research~\citep{dontdecay,largebatch-eval-sys} and has been shown particularly useful for training the Transformer architecture with the Tensor2Tensor framework~\citep{t2t-training}. We support these findings in Tab.~\ref{tab:batch-size}.\footnote{We had to reduce the learning rate for $g'=1$ to avoid training divergence.} Our technical infrastructure\footnote{\url{http://www.hpc.cam.ac.uk/}} allows us to train on four P100 GPUs simultaneously, which limits the number of physical GPUs to $g=4$ and the batch size\footnote{We follow \citet{transformer,t2t} and specify the batch size in terms of number of source and target tokens in a batch, not the number of sentences.} to $b=2048$ due to the GPU memory. Thus, the maximum possible effective batch size without delaying SGD updates is $b'=8192$. Training with delay factor $d$ accumulates gradients over $d$ batches and applies the optimizer update rule on the accumulated gradients. This allows us to scale up the effective number of GPUs to 16 and improve the BLEU score significantly (29.5 vs.\ 30.3). Note that training regimens are equivalent if their effective batch size is the same, ie.\ training on 4 physical GPUs with $d=4$ is mathematically equivalent to training on 16 GPUs without delaying SGD updates.  Tab.~\ref{tab:training} lists our training setups for the neural architectures used in this work. These training hyper-parameters were chosen empirically. Particularly, we did not find improvements by increasing the number of effective GPUs for SliceNet or longer LSTM training.

We use {\em news-test2017} as development set on all language pairs to tune the model interpolation weights $\lambda$ (Eq.~\ref{eq:combined-score}) and the scaling factor for length normalization.

\subsection{Decoding}

We use the \texttt{beam} search strategy with beam size 8 of the SGNMT decoder~\citep{sgnmt1,sgnmt2} in all our experiments. We apply length normalization~\citep{bahdanau} on German-English and Chinese-English but not on English-German. As outlined in Sec.~\ref{sec:sys-combination} we either use full posteriors or MBR-style $n$-gram posteriors from our individual models. SMT $n$-gram scores are extracted as described by \citet{lmbr-hifst} using HiFST's \texttt{lmbr} tool. We use SGNMT's \texttt{ngram} output format to extract $n$-gram scores from our neural models.

\section{Results}

On English-German and German-English {\em news-test2014} we compute cased BLEU scores with Moses' \texttt{multi-bleu.pl} script on tokenized output to be comparable with prior work~\citep{gnmt,slicenet,convs2s,transformer,rnmt}. On all other test sets we use \texttt{mteval-v13a.pl} to be comparable to the official cased WMT scores.\footnote{\url{http://matrix.statmt.org/}}

\begin{table*}[t!]
\begin{center}
\small
\begin{tabular}{|@{\hspace{0.3em}}l@{\hspace{0.3em}}|@{\hspace{0.3em}}c@{\hspace{0.3em}}|c|c|c|c|c|c|c|c|c|c|}
\hline
\textbf{Architecture} & \textbf{\#Systems} & \multicolumn{4}{c|}{\textbf{English-German}} & \multicolumn{4}{c|}{\textbf{German-English}} & \multicolumn{2}{c|}{\textbf{Chinese-English}} \\
 &  & test14 & test15 & test16 & test17 & test14 & test15 & test16 & test17 & dev17 & test17\\
\hline
PBMT & 1 & 19.6 & 20.9 & 25.6 & 20.0 & 22.5 & 27.2 & 32.6 & 28.2 & 14.2 & 15.8 \\
\hline
\multirow{2}{*}{LSTM} & 1 & 27.1 & 28.8 & 34.6 & 28.0 & 33.8 & 33.3 & 40.7 & 34.8 & 21.8 & 22.7 \\
 & 2 & 28.2 & 29.6 & 35.5 & 28.5 & 34.6 & 34.0 & 41.4 & 35.3 & 22.7 & 23.6 \\
\hline
\multirow{2}{*}{SliceNet} & 1 & 26.8 & 28.9 & 33.6 & 27.6 & 32.6 & 32.3 & 39.8 & 33.7 & 21.4 & 22.5 \\
 & 2 & 27.2 & 29.6 & 34.6 & 28.3 & 33.2 & 32.9 & 40.8 & 34.3 & 21.8 & 23.4 \\
\hline
R2L Trans. & 1 & 30.3 & 31.5 & 36.3 & 30.2 & 36.5 & 35.5 & 43.5 & 37.2 & 24.5 & 24.9 \\
\hline
\multirow{2}{*}{Transformer} & 1 & 30.7 & 31.9 & 36.6 & 30.5 & 36.7 & 36.2 & 43.7 & 37.9 & 24.9 & 25.6 \\
 & 2 & 31.1 & 31.8 & 37.2 & 31.0 & 36.9	 & 36.4 & 44.0 & 38.1 & 26.2 & 26.2 \\
\hline
\multirow{2}{*}{Rel.\ Trans.} & 1 & 31.2 & 31.9 & 37.0 & 31.1 & 37.0 & 36.3 & 44.1 & 38.1 & 24.9 & 25.8 \\
 & 2 & 31.4 & 32.3 & 37.7 & 31.2 & 37.2 & 36.5 & 44.1 & 38.4 & 25.1 & 26.4 \\
\hline
\end{tabular}
\end{center}
\caption{\label{tab:single-architecture} Single architecture results on all language pairs for single systems and 2-ensembles.}
\end{table*}

\begin{table*}[t!]
\begin{center}
\small
\begin{tabular}{@{\hspace{0em}}r@{\hspace{0.2em}}|c@{\hspace{0.3em}}|@{\hspace{0.3em}}c@{\hspace{0.3em}}|@{\hspace{0.3em}}c@{\hspace{0.3em}}|@{\hspace{0.3em}}c@{\hspace{0.3em}}|@{\hspace{0.3em}}c||c@{\hspace{0.3em}}|@{\hspace{0.3em}}c@{\hspace{0.3em}}|@{\hspace{0.3em}}c@{\hspace{0.3em}}|@{\hspace{0.3em}}c||c|c|c|}
\cline{2-13}
& \multicolumn{5}{c||}{\textbf{Full posterior}} & \multicolumn{4}{c||}{\textbf{MBR-based $n$-gram scores}} & \multicolumn{3}{c|}{\textbf{BLEU (test2017)}} \\
 & PBMT & LSTM$^*$ & SliceNet$^*$ & Trans. & Rel.\ Trans. & PBMT & LSTM$^*$ & SliceNet$^*$ & R2L Trans. & en-de & de-en & zh-en \\
\cline{2-13}
\footnotesize{1} & $\checkmark$ &  &  &  &  &  & &  & & 20.0 & 28.2 & 15.8 \\
\footnotesize{2} & & $\checkmark$ &  &  &  &  & &  & & 28.5 & 35.3 & 23.6 \\
\footnotesize{3} & & & $\checkmark$ &  &  &  & &  & & 28.3 & 34.3 & 23.4 \\
\footnotesize{4} & & &  & $\checkmark$ &  &  & &  & & 30.5 & 37.9 & 25.6 \\
\footnotesize{5} & & &  &  & $\checkmark$ &  & &  & & 31.1 & 38.1 & 25.8 \\
\cline{2-13}
\footnotesize{6} & & &  & $\checkmark$ & $\checkmark$ &  & &  & & 31.3 & 38.2 & 26.4 \\
\footnotesize{7} & & $\checkmark$ & $\checkmark$ & $\checkmark$ & $\checkmark$ &  & &  & & 31.3 & 38.2 & 26.4 \\
\cline{2-13}
\footnotesize{8} & &  &  & $\checkmark$ & $\checkmark$ & & $\checkmark$ & $\checkmark$ & & 31.4 & 38.2 & 26.6 \\
\footnotesize{9} & &  &  & $\checkmark$ & $\checkmark$ & & $\checkmark$ & $\checkmark$ & $\checkmark$ & 31.4 & 38.3 & 26.8 \\
\footnotesize{10} & &  &  & $\checkmark$ & $\checkmark$ & $\checkmark$ & $\checkmark$ & $\checkmark$ & $\checkmark$ & 31.7 & 38.7 & 27.1 \\
\cline{2-13}
\end{tabular}
\end{center}
\caption{\label{tab:model-combination} Model combination with ensembling and MBR.Model scores are weighted with MERT and combined (log-)linearly as described in Sec.~\ref{sec:sys-combination}. $^*$:  The LSTM and SliceNet models are 2-ensembles.}
\end{table*}

First, we will discuss our experiments with a single architecture, i.e.\ single systems and ensembles of two systems with the same architecture. Tab.~\ref{tab:single-architecture} compares the architectures on all test sets. PBMT as a single system is clearly inferior to all neural systems. Ensembling neural systems helps for all architectures across the board. LSTM is usually slightly better than the convolutional SliceNet, but is much slower to train and decode (cf.\ Tab.~\ref{tab:params}). Note that our LSTM 2-ensemble is on par with the best BLEU score in WMT17~\citep{uedin-wmt17}, which was also based on recurrent models. Transformer architectures outperform LSTMs and SliceNets on all test sets. The right-to-left Transformer is usually slightly worse, the Transformer with relative positioning slightly better than the standard Transformer setup.

Tab.~\ref{tab:model-combination} summarizes our system combination results with multiple architectures. Adding LSTM and SliceNet as full-posterior models to an ensemble of a Transformer and a Relative Transformer does not improve the BLEU score (rows 6 vs.\ 7). We see very slight improvements when we use these models to extract $n$-gram scores instead (rows 6 vs.\ 8). We report further gains by using MBR-based $n$-gram scores from the right-to-left Transformer and the PBMT system. The improvements from adding PBMT are rather small, but we still found them surprising given that the PBMT baseline is usally more than 10 BLEU points worse than our best single neural model. We list the performance of our submitted systems on all test sets in Tab.~\ref{tab:final-results}.

\begin{table}[t!]
\begin{center}
\small
\begin{tabular}{|l|c|r|}
\hline
\textbf{Direction} & \textbf{Test set} & \textbf{BLEU} \\
\hline
\multirow{5}{*}{English-German} & news-test14 & 31.6 \\
 & news-test15 & 32.6 \\
 & news-test16 & 38.5 \\
 & news-test17 & 31.7 \\
 & news-test18 & 46.6 \\
\hline
\multirow{5}{*}{German-English} & news-test14 & 36.8 \\
 & news-test15 & 36.5 \\
 & news-test16 & 45.1 \\
 & news-test17 & 38.7 \\
 & news-test18 & 48.0 \\
\hline
\multirow{3}{*}{Chinese-English} & news-dev17 & 25.7 \\
 & news-test17 & 27.1 \\
 & news-test18 & 27.7 \\
\hline
\end{tabular}
\end{center}
\caption{\label{tab:final-results} BLEU scores of the submitted systems (row 10 in Tab.~\ref{tab:model-combination}).}
\end{table}

\section{Related Work}

There is a large body of research comparing NMT and SMT~\citep{nmt-vs-smt-monotone,nmt-vs-smt-9-langs,nmt-vs-smt-challenges,nmt-vs-smt-dead,nmt-vs-smt-irish,nmt-vs-smt-case-bentivogli1,nmt-vs-smt-case-bentivogli2}. Most studies have found superior overall translation quality of NMT models in most settings, but complementary strengths of both paradigms. Therefore, the literature about hybrid NMT-SMT systems is also vast, ranging from rescoring and reranking methods~\citep{hybrid-neubig-nbest,hybrid-sgnmt,hybrid-rescoring-adaptation,hybrid-gec-rescoring,hybrid-reranking-german,hybrid-smorgasbord}, MBR-based formalisms~\citep{mbr-nmt,sgnmt2,hybrid-mbr-sdl}, NMT assisting SMT~\citep{hybrid-nmt-features,hybrid-neural-pre-translation}, and SMT assisting NMT~\citep{hybrid-pre-translation,hybrid-smt-features,hybrid-technical-terms,hybrid-smt-advise,hybrid-smt-features-search,hybrid-nmt-multisource}. We confirm the potential of hybrid systems by reporting gains on top of very strong neural ensembles.

Ensembling is a well-known technique in NMT to improve system performance. However, ensembles usually consist of multiple models of the same architecture. In this paper, we compare and combine three very different architectures (recurrent, convolutional, and self-attention based) in two different ways (full posterior and MBR-based), and find that combination with MBR-based $n$-gram scores is superior.

\section{Conclusion}

We have described our WMT18 submission, which achieves very competitive BLEU scores on all three language pairs (English-German, German-English, and Chinese-English) and significantly higher accuracies in a variety of linguistic phenomena compared to other submissions~\citep{wmt18-deen-ana}. Our system combines three different neural architecture with a traditional PBMT system. We showed that our MBR-based scheme is effective to combine these diverse models of translation, and that adding the PBMT system to the mix of neural models still yields gains although it is much worse as stand-alone system.

\section*{Acknowledgments}

This work was supported in part by the U.K. Engineering and Physical Sciences Research Council (EPSRC grant EP/L027623/1).


\bibliography{refs}

\begin{thebibliography}{67}
\expandafter\ifx\csname natexlab\endcsname\relax\def\natexlab#1{#1}\fi

\bibitem[{Ahmed et~al.(2017)Ahmed, Keskar, and Socher}]{weighted-transformer}
Karim Ahmed, Nitish~Shirish Keskar, and Richard Socher. 2017.
\newblock Weighted transformer network for machine translation.
\newblock \emph{arXiv preprint arXiv:1711.02132}.

\bibitem[{Avramidis et~al.(2016)Avramidis, Macketanz, Burchardt, Helcl, and
  Uszkoreit}]{hybrid-reranking-german}
Eleftherios Avramidis, Vivien Macketanz, Aljoscha Burchardt, Jindrich Helcl,
  and Hans Uszkoreit. 2016.
\newblock Deeper machine translation and evaluation for {German}.
\newblock In \emph{Proceedings of the 2nd Deep Machine Translation Workshop},
  pages 29--38. {\'U}FAL MFF UK.

\bibitem[{Avramidis et~al.(2018)}]{wmt18-deen-ana}
Eleftherios Avramidis et~al. 2018.
\newblock Fine-grained evaluation of {German-English} machine translation based
  on a test suite.
\newblock In \emph{Proceedings of the Third Conference on Machine Translation,
  Volume 3}, Brussels, Belgium. Association for Computational Linguistics.

\bibitem[{Ba et~al.(2016)Ba, Kiros, and Hinton}]{layernorm}
Jimmy~Lei Ba, Jamie~Ryan Kiros, and Geoffrey~E Hinton. 2016.
\newblock Layer normalization.
\newblock \emph{arXiv preprint arXiv:1607.06450}.

\bibitem[{Bahdanau et~al.(2015)Bahdanau, Cho, and Bengio}]{bahdanau}
Dzmitry Bahdanau, Kyunghyun Cho, and Yoshua Bengio. 2015.
\newblock Neural machine translation by jointly learning to align and
  translate.
\newblock In \emph{Proceedings of the International Conference on Learning
  Representations (ICLR)}, Toulon, France.

\bibitem[{Bentivogli et~al.(2016)Bentivogli, Bisazza, Cettolo, and
  Federico}]{nmt-vs-smt-case-bentivogli1}
Luisa Bentivogli, Arianna Bisazza, Mauro Cettolo, and Marcello Federico. 2016.
\newblock Neural versus phrase-based machine translation quality: a case study.
\newblock In \emph{Proceedings of the 2016 Conference on Empirical Methods in
  Natural Language Processing}, pages 257--267. Association for Computational
  Linguistics.

\bibitem[{Bentivogli et~al.(2018)Bentivogli, Bisazza, Cettolo, and
  Federico}]{nmt-vs-smt-case-bentivogli2}
Luisa Bentivogli, Arianna Bisazza, Mauro Cettolo, and Marcello Federico. 2018.
\newblock {Neural versus phrase-based MT quality: An in-depth analysis on
  English–German and English–French}.
\newblock \emph{Computer Speech \& Language}, 49:52--70.

\bibitem[{Blackwood et~al.(2010)Blackwood, Gispert, and Byrne}]{lmbr-hifst}
Graeme Blackwood, Adri{\`a} Gispert, and William Byrne. 2010.
\newblock Efficient path counting transducers for minimum {Bayes}-risk decoding
  of statistical machine translation lattices.
\newblock In \emph{Proceedings of the ACL 2010 Conference Short Papers}, pages
  27--32. Association for Computational Linguistics.

\bibitem[{Chen et~al.(2018)Chen, Firat, Bapna, Johnson, Macherey, Foster,
  Jones, Parmar, Schuster, Chen et~al.}]{rnmt}
Mia~Xu Chen, Orhan Firat, Ankur Bapna, Melvin Johnson, Wolfgang Macherey,
  George Foster, Llion Jones, Niki Parmar, Mike Schuster, Zhifeng Chen, et~al.
  2018.
\newblock The best of both worlds: Combining recent advances in neural machine
  translation.
\newblock \emph{arXiv preprint arXiv:1804.09849}.

\bibitem[{Chrisman(1991)}]{early-raam2}
Lonnie Chrisman. 1991.
\newblock Learning recursive distributed representations for holistic
  computation.
\newblock \emph{Connection Science}, 3(4):345--366.

\bibitem[{Dahlmann et~al.(2017)Dahlmann, Matusov, Petrushkov, and
  Khadivi}]{hybrid-smt-features-search}
Leonard Dahlmann, Evgeny Matusov, Pavel Petrushkov, and Shahram Khadivi. 2017.
\newblock Neural machine translation leveraging phrase-based models in a hybrid
  search.
\newblock In \emph{Proceedings of the 2017 Conference on Empirical Methods in
  Natural Language Processing}, pages 1411--1420. Association for Computational
  Linguistics.

\bibitem[{Dowling et~al.(2018)Dowling, Lynn, Poncelas, and
  Way}]{nmt-vs-smt-irish}
Meghan Dowling, Teresa Lynn, Alberto Poncelas, and Andy Way. 2018.
\newblock {SMT versus NMT: Preliminary comparisons for Irish}.
\newblock \emph{Technologies for MT of Low Resource Languages (LoResMT 2018)},
  page~12.

\bibitem[{Du and Way(2017)}]{hybrid-neural-pre-translation}
Jinhua Du and Andy Way. 2017.
\newblock Neural pre-translation for hybrid machine translation.
\newblock \emph{In Proceedings of MT Summit XVI}, 1:27--40.

\bibitem[{Forcada and {\~{N}}eco(1997)}]{early-raam3}
Mikel~L. Forcada and Ram{\'o}n~P. {\~{N}}eco. 1997.
\newblock Recursive hetero-associative memories for translation.
\newblock In \emph{Biological and Artificial Computation: From Neuroscience to
  Technology}, pages 453--462, Berlin, Heidelberg. Springer Berlin Heidelberg.

\bibitem[{Gehring et~al.(2017)Gehring, Auli, Grangier, Yarats, and
  Dauphin}]{convs2s}
Jonas Gehring, Michael Auli, David Grangier, Denis Yarats, and Yann~N Dauphin.
  2017.
\newblock Convolutional sequence to sequence learning.
\newblock \emph{ArXiv e-prints}.

\bibitem[{Grundkiewicz and Junczys-Dowmunt(2018)}]{hybrid-gec-rescoring}
Roman Grundkiewicz and Marcin Junczys-Dowmunt. 2018.
\newblock Near human-level performance in grammatical error correction with
  hybrid machine translation.
\newblock In \emph{Proceedings of the 2018 Conference of the North American
  Chapter of the Association for Computational Linguistics: Human Language
  Technologies}. Association for Computational Linguistics.

\bibitem[{Hassan et~al.(2018)Hassan, Aue, Chen, Chowdhary, Clark, Federmann,
  Huang, Junczys-Dowmunt, Lewis, Li et~al.}]{microsoft-parity}
Hany Hassan, Anthony Aue, Chang Chen, Vishal Chowdhary, Jonathan Clark,
  Christian Federmann, Xuedong Huang, Marcin Junczys-Dowmunt, William Lewis,
  Mu~Li, et~al. 2018.
\newblock Achieving human parity on automatic {Chinese to English} news
  translation.
\newblock \emph{arXiv preprint arXiv:1803.05567}.

\bibitem[{He et~al.(2016)He, He, Wu, and Wang}]{hybrid-smt-features}
Wei He, Zhongjun He, Hua Wu, and Haifeng Wang. 2016.
\newblock Improved neural machine translation with {SMT} features.
\newblock In \emph{AAAI}, pages 151--157.

\bibitem[{Iglesias et~al.(2018)Iglesias, Tambellini, de~Gispert, Hasler, and
  Byrne}]{hybrid-mbr-sdl}
Gonzalo Iglesias, William Tambellini, Adri{\`a} de~Gispert, Eva Hasler, and
  Bill Byrne. 2018.
\newblock Accelerating {NMT} batched beam decoding with {LMBR} posteriors for
  deployment.
\newblock In \emph{Proceedings of the 2018 Conference of the North American
  Chapter of the Association for Computational Linguistics: Human Language
  Technologies}. Association for Computational Linguistics.

\bibitem[{Junczys-Dowmunt et~al.(2016{\natexlab{a}})Junczys-Dowmunt, Dwojak,
  and Hoang}]{marian}
Marcin Junczys-Dowmunt, Tomasz Dwojak, and Hieu Hoang. 2016{\natexlab{a}}.
\newblock Is neural machine translation ready for deployment? {A} case study on
  30 translation directions.
\newblock In \emph{Proceedings of the 9th International Workshop on Spoken
  Language Translation (IWSLT)}, Seattle, WA.

\bibitem[{Junczys-Dowmunt et~al.(2016{\natexlab{b}})Junczys-Dowmunt, Dwojak,
  and Sennrich}]{hybrid-nmt-features}
Marcin Junczys-Dowmunt, Tomasz Dwojak, and Rico Sennrich. 2016{\natexlab{b}}.
\newblock The {AMU-UEDIN} submission to the {WMT16} news translation task:
  Attention-based {NMT} models as feature functions in phrase-based smt.
\newblock In \emph{Proceedings of the First Conference on Machine Translation:
  Volume 2, Shared Task Papers}, pages 319--325. Association for Computational
  Linguistics.

\bibitem[{Kaiser et~al.(2017)Kaiser, Gomez, and Chollet}]{slicenet}
Lukasz Kaiser, Aidan~N Gomez, and Francois Chollet. 2017.
\newblock Depthwise separable convolutions for neural machine translation.
\newblock \emph{arXiv preprint arXiv:1706.03059}.

\bibitem[{Kalchbrenner and Blunsom(2013)}]{encdec}
Nal Kalchbrenner and Phil Blunsom. 2013.
\newblock Recurrent continuous translation models.
\newblock In \emph{Proceedings of the 2013 Conference on Empirical Methods in
  Natural Language Processing}, pages 1700--1709, Seattle, Washington, USA.
  Association for Computational Linguistics.

\bibitem[{Khayrallah et~al.(2017)Khayrallah, Kumar, Duh, Post, and
  Koehn}]{hybrid-rescoring-adaptation}
Huda Khayrallah, Gaurav Kumar, Kevin Duh, Matt Post, and Philipp Koehn. 2017.
\newblock Neural lattice search for domain adaptation in machine translation.
\newblock In \emph{Proceedings of the Eighth International Joint Conference on
  Natural Language Processing (Volume 2: Short Papers)}, pages 20--25. Asian
  Federation of Natural Language Processing.

\bibitem[{Kiefer(1953)}]{golden-section-search}
Jack Kiefer. 1953.
\newblock Sequential minimax search for a maximum.
\newblock \emph{Proceedings of the American mathematical society},
  4(3):502--506.

\bibitem[{Kingma and Ba(2015)}]{adam}
Diederik~P Kingma and Jimmy Ba. 2015.
\newblock Adam: A method for stochastic optimization.
\newblock In \emph{Proceedings of the International Conference on Learning
  Representations (ICLR)}.

\bibitem[{Koehn and Knowles(2017)}]{nmt-vs-smt-challenges}
Philipp Koehn and Rebecca Knowles. 2017.
\newblock Six challenges for neural machine translation.
\newblock In \emph{Proceedings of the First Workshop on Neural Machine
  Translation}, pages 28--39. Association for Computational Linguistics.

\bibitem[{Kumar and Byrne(2004)}]{mbr-shankar}
Shankar Kumar and William Byrne. 2004.
\newblock Minimum {Bayes}-risk decoding for statistical machine translation.
\newblock In \emph{Proceedings of the Human Language Technology Conference of
  the North American Chapter of the Association for Computational Linguistics:
  HLT-NAACL 2004}.

\bibitem[{Li et~al.(2017)Li, Zhang, Wang, and Zheng}]{right-to-left-draft}
Aodong Li, Shiyue Zhang, Dong Wang, and Thomas~Fang Zheng. 2017.
\newblock Enhanced neural machine translation by learning from draft.
\newblock In \emph{Proceedings of APSIPA Annual Summit and Conference}, volume
  2017, pages 12--15.

\bibitem[{Li et~al.(2013)Li, Liu, and Sun}]{li-recursive-autoencoders}
Peng Li, Yang Liu, and Maosong Sun. 2013.
\newblock Recursive autoencoders for {ITG}-based translation.
\newblock In \emph{Proceedings of the 2013 Conference on Empirical Methods in
  Natural Language Processing}, pages 567--577, Seattle, Washington, USA.
  Association for Computational Linguistics.

\bibitem[{Liu et~al.(2016)Liu, Utiyama, Finch, and
  Sumita}]{right-to-left-agreement}
Lemao Liu, Masao Utiyama, Andrew Finch, and Eiichiro Sumita. 2016.
\newblock Agreement on target-bidirectional neural machine translation.
\newblock In \emph{Proceedings of the 2016 Conference of the North American
  Chapter of the Association for Computational Linguistics: Human Language
  Technologies}, pages 411--416. Association for Computational Linguistics.

\bibitem[{Long et~al.(2016)Long, Utsuro, Mitsuhashi, and
  Yamamoto}]{hybrid-technical-terms}
Zi~Long, Takehito Utsuro, Tomoharu Mitsuhashi, and Mikio Yamamoto. 2016.
\newblock Translation of patent sentences with a large vocabulary of technical
  terms using neural machine translation.
\newblock In \emph{Proceedings of the 3rd Workshop on Asian Translation
  (WAT2016)}, pages 47--57. The COLING 2016 Organizing Committee.

\bibitem[{Luong et~al.(2017)Luong, Brevdo, and Zhao}]{tfnmt}
Minh{-}Thang Luong, Eugene Brevdo, and Rui Zhao. 2017.
\newblock Neural machine translation (seq2seq) tutorial.
\newblock \emph{https://github.com/tensorflow/nmt}.

\bibitem[{Luong et~al.(2015)Luong, Pham, and Manning}]{dot-product-attention}
Thang Luong, Hieu Pham, and Christopher~D. Manning. 2015.
\newblock Effective approaches to attention-based neural machine translation.
\newblock In \emph{Proceedings of the 2015 Conference on Empirical Methods in
  Natural Language Processing}, pages 1412--1421, Lisbon, Portugal. Association
  for Computational Linguistics.

\bibitem[{Macherey et~al.(2008)Macherey, Och, Thayer, and Uszkoreit}]{lmert}
Wolfgang Macherey, Franz Och, Ignacio Thayer, and Jakob Uszkoreit. 2008.
\newblock Lattice-based minimum error rate training for statistical machine
  translation.
\newblock In \emph{Proceedings of the 2008 Conference on Empirical Methods in
  Natural Language Processing}, pages 725--734. Association for Computational
  Linguistics.

\bibitem[{Marie and Fujita(2018)}]{hybrid-smorgasbord}
Benjamin Marie and Atsushi Fujita. 2018.
\newblock A smorgasbord of features to combine phrase -based and neural machine
  translation.
\newblock In \emph{Proceedings of AMTA Workshop on MT Research and the
  Translation Industry}, Boston, US.

\bibitem[{Menacer et~al.(2017)Menacer, Langlois, Mella, Fohr, Jouvet, and
  Sma{\"\i}li}]{nmt-vs-smt-dead}
Mohamed-Amine Menacer, David Langlois, Odile Mella, Dominique Fohr, Denis
  Jouvet, and Kamel Sma{\"\i}li. 2017.
\newblock Is statistical machine translation approach dead?
\newblock In \emph{ICNLSSP 2017-International Conference on Natural Language,
  Signal and Speech Processing}.

\bibitem[{Nakatani(2010)}]{langdetect}
Shuyo Nakatani. 2010.
\newblock Language detection library for {Java}.

\bibitem[{Neishi et~al.(2017)Neishi, Sakuma, Tohda, Ishiwatari, Yoshinaga, and
  Toyoda}]{largebatch-eval-sys}
Masato Neishi, Jin Sakuma, Satoshi Tohda, Shonosuke Ishiwatari, Naoki
  Yoshinaga, and Masashi Toyoda. 2017.
\newblock A bag of useful tricks for practical neural machine translation:
  Embedding layer initialization and large batch size.
\newblock In \emph{Proceedings of the 4th Workshop on Asian Translation
  (WAT2017)}, pages 99--109.

\bibitem[{Neubig et~al.(2015)Neubig, Morishita, and
  Nakamura}]{hybrid-neubig-nbest}
Graham Neubig, Makoto Morishita, and Satoshi Nakamura. 2015.
\newblock Neural reranking improves subjective quality of machine translation:
  {NAIST} at {WAT}2015.
\newblock In \emph{WAT}, Kyoto, Japan.

\bibitem[{Niehues et~al.(2016)Niehues, Cho, Ha, and
  Waibel}]{hybrid-pre-translation}
Jan Niehues, Eunah Cho, Thanh-Le Ha, and Alex Waibel. 2016.
\newblock Pre-translation for neural machine translation.
\newblock In \emph{Proceedings of COLING 2016, the 26th International
  Conference on Computational Linguistics: Technical Papers}, pages 1828--1836.
  The COLING 2016 Organizing Committee.

\bibitem[{Och(2003)}]{nbest-mert}
Franz~Josef Och. 2003.
\newblock Minimum error rate training in statistical machine translation.
\newblock In \emph{Proceedings of the 41st Annual Meeting of the Association
  for Computational Linguistics}.

\bibitem[{Pollack(1990)}]{early-raam1}
Jordan~B. Pollack. 1990.
\newblock Recursive distributed representations.
\newblock \emph{Artificial Intelligence}, 46(1):77 -- 105.

\bibitem[{Popel and Bojar(2018)}]{t2t-training}
Martin Popel and Ond{\v{r}}ej Bojar. 2018.
\newblock Training tips for the {Transformer} model.
\newblock \emph{arXiv preprint arXiv:1804.00247}.

\bibitem[{Powell(1964)}]{powell}
Michael~JD Powell. 1964.
\newblock An efficient method for finding the minimum of a function of several
  variables without calculating derivatives.
\newblock \emph{The computer journal}, 7(2):155--162.

\bibitem[{Saunders et~al.(2018)Saunders, Stahlberg, de~Gispert, and
  Byrne}]{danielle-syntax}
Danielle Saunders, Felix Stahlberg, Adri{\`a} de~Gispert, and Bill Byrne. 2018.
\newblock Multi-representation ensembles and delayed {SGD} updates improve
  syntax-based {NMT}.
\newblock In \emph{Proceedings of the 56th Annual Meeting of the Association
  for Computational Linguistics}. Association for Computational Linguistics.
\newblock To appear.

\bibitem[{Schnober et~al.(2016)Schnober, Eger, Do~Dinh, and
  Gurevych}]{nmt-vs-smt-monotone}
Carsten Schnober, Steffen Eger, Erik-L{\^a}n Do~Dinh, and Iryna Gurevych. 2016.
\newblock Still not there? {Comparing} traditional sequence-to-sequence models
  to encoder-decoder neural networks on monotone string translation tasks.
\newblock In \emph{Proceedings of COLING 2016, the 26th International
  Conference on Computational Linguistics: Technical Papers}, pages 1703--1714.
  The COLING 2016 Organizing Committee.

\bibitem[{Sennrich et~al.(2017)Sennrich, Birch, Currey, Germann, Haddow,
  Heafield, Miceli~Barone, and Williams}]{uedin-wmt17}
Rico Sennrich, Alexandra Birch, Anna Currey, Ulrich Germann, Barry Haddow,
  Kenneth Heafield, Antonio~Valerio Miceli~Barone, and Philip Williams. 2017.
\newblock The {University of Edinburgh's} neural {MT} systems for {WMT17}.
\newblock In \emph{Proceedings of the Second Conference on Machine Translation,
  Volume 2: Shared Task Papers}, pages 389--399, Copenhagen, Denmark.
  Association for Computational Linguistics.

\bibitem[{Sennrich et~al.(2016{\natexlab{a}})Sennrich, Haddow, and
  Birch}]{backtranslation}
Rico Sennrich, Barry Haddow, and Alexandra Birch. 2016{\natexlab{a}}.
\newblock Improving neural machine translation models with monolingual data.
\newblock In \emph{Proceedings of the 54th Annual Meeting of the Association
  for Computational Linguistics (Volume 1: Long Papers)}, pages 86--96.
  Association for Computational Linguistics.

\bibitem[{Sennrich et~al.(2016{\natexlab{b}})Sennrich, Haddow, and Birch}]{bpe}
Rico Sennrich, Barry Haddow, and Alexandra Birch. 2016{\natexlab{b}}.
\newblock Neural machine translation of rare words with subword units.
\newblock In \emph{Proceedings of the 54th Annual Meeting of the Association
  for Computational Linguistics (Volume 1: Long Papers)}, pages 1715--1725.
  Association for Computational Linguistics.

\bibitem[{Shaw et~al.(2018)Shaw, Uszkoreit, and Vaswani}]{relative-transformer}
Peter Shaw, Jakob Uszkoreit, and Ashish Vaswani. 2018.
\newblock Self-attention with relative position representations.
\newblock In \emph{Proceedings of the 2018 Conference of the North American
  Chapter of the Association for Computational Linguistics: Human Language
  Technologies}. Association for Computational Linguistics.

\bibitem[{Smith et~al.(2017)Smith, Kindermans, and Le}]{dontdecay}
Samuel~L Smith, Pieter-Jan Kindermans, and Quoc~V Le. 2017.
\newblock Don't decay the learning rate, increase the batch size.
\newblock \emph{arXiv preprint arXiv:1711.00489}.

\bibitem[{Socher et~al.(2011)Socher, Pennington, Huang, Ng, and
  Manning}]{socher-recursive-autoencoders}
Richard Socher, Jeffrey Pennington, Eric~H. Huang, Andrew~Y. Ng, and
  Christopher~D. Manning. 2011.
\newblock Semi-supervised recursive autoencoders for predicting sentiment
  distributions.
\newblock In \emph{Proceedings of the 2011 Conference on Empirical Methods in
  Natural Language Processing}, pages 151--161, Edinburgh, Scotland, UK.
  Association for Computational Linguistics.

\bibitem[{Srivastava et~al.(2014)Srivastava, Hinton, Krizhevsky, Sutskever, and
  Salakhutdinov}]{dropout}
Nitish Srivastava, Geoffrey Hinton, Alex Krizhevsky, Ilya Sutskever, and Ruslan
  Salakhutdinov. 2014.
\newblock Dropout: A simple way to prevent neural networks from overfitting.
\newblock \emph{The Journal of Machine Learning Research}, 15(1):1929--1958.

\bibitem[{Stahlberg et~al.(2017{\natexlab{a}})Stahlberg, de~Gispert, Hasler,
  and Byrne}]{mbr-nmt}
Felix Stahlberg, Adri{\`a} de~Gispert, Eva Hasler, and Bill Byrne.
  2017{\natexlab{a}}.
\newblock Neural machine translation by minimising the {Bayes}-risk with
  respect to syntactic translation lattices.
\newblock In \emph{Proceedings of the 15th Conference of the European Chapter
  of the Association for Computational Linguistics: Volume 2, Short Papers},
  pages 362--368. Association for Computational Linguistics.

\bibitem[{Stahlberg et~al.(2017{\natexlab{b}})Stahlberg, Hasler, Saunders, and
  Byrne}]{sgnmt1}
Felix Stahlberg, Eva Hasler, Danielle Saunders, and Bill Byrne.
  2017{\natexlab{b}}.
\newblock {SGNMT -- A} flexible {NMT} decoding platform for quick prototyping
  of new models and search strategies.
\newblock In \emph{Proceedings of the 2017 Conference on Empirical Methods in
  Natural Language Processing: System Demonstrations}, pages 25--30.
  Association for Computational Linguistics.

\bibitem[{Stahlberg et~al.(2016)Stahlberg, Hasler, Waite, and
  Byrne}]{hybrid-sgnmt}
Felix Stahlberg, Eva Hasler, Aurelien Waite, and Bill Byrne. 2016.
\newblock Syntactically guided neural machine translation.
\newblock In \emph{Proceedings of the 54th Annual Meeting of the Association
  for Computational Linguistics (Volume 2: Short Papers)}, pages 299--305.
  Association for Computational Linguistics.

\bibitem[{Stahlberg et~al.(2018)Stahlberg, Saunders, Iglesias, and
  Byrne}]{sgnmt2}
Felix Stahlberg, Danielle Saunders, Gonzalo Iglesias, and Bill Byrne. 2018.
\newblock Why not be versatile? {Applications} of the {SGNMT} decoder for
  machine translation.
\newblock In \emph{Proceedings of AMTA Workshop on MT Research and the
  Translation Industry}, Boston, US.

\bibitem[{Sutskever et~al.(2014)Sutskever, Vinyals, and Le}]{sutskever}
Ilya Sutskever, Oriol Vinyals, and Quoc~V Le. 2014.
\newblock Sequence to sequence learning with neural networks.
\newblock In \emph{Advances in Neural Information Processing Systems 27}, pages
  3104--3112. Curran Associates, Inc.

\bibitem[{Szegedy et~al.(2016)Szegedy, Vanhoucke, Ioffe, Shlens, and
  Wojna}]{label-smoothing}
Christian Szegedy, Vincent Vanhoucke, Sergey Ioffe, Jon Shlens, and Zbigniew
  Wojna. 2016.
\newblock Rethinking the inception architecture for computer vision.
\newblock In \emph{Proceedings of the IEEE Conference on Computer Vision and
  Pattern Recognition}, pages 2818--2826.

\bibitem[{Toral and S{\'a}nchez-Cartagena(2017)}]{nmt-vs-smt-9-langs}
Antonio Toral and V{\'i}ctor~M. S{\'a}nchez-Cartagena. 2017.
\newblock A multifaceted evaluation of neural versus phrase-based machine
  translation for 9 language directions.
\newblock In \emph{Proceedings of the 15th Conference of the European Chapter
  of the Association for Computational Linguistics: Volume 1, Long Papers},
  pages 1063--1073. Association for Computational Linguistics.

\bibitem[{Tromble et~al.(2008)Tromble, Kumar, Och, and Macherey}]{mbr-tromble}
Roy Tromble, Shankar Kumar, Franz Och, and Wolfgang Macherey. 2008.
\newblock Lattice minimum {Bayes}-risk decoding for statistical machine
  translation.
\newblock In \emph{Proceedings of the 2008 Conference on Empirical Methods in
  Natural Language Processing}, pages 620--629. Association for Computational
  Linguistics.

\bibitem[{Vaswani et~al.(2018)Vaswani, Bengio, Brevdo, Chollet, Gomez, Gouws,
  Jones, Kaiser, Kalchbrenner, Parmar et~al.}]{t2t}
Ashish Vaswani, Samy Bengio, Eugene Brevdo, Francois Chollet, Aidan~N Gomez,
  Stephan Gouws, Llion Jones, {\L}ukasz Kaiser, Nal Kalchbrenner, Niki Parmar,
  et~al. 2018.
\newblock Tensor2tensor for neural machine translation.
\newblock In \emph{Proceedings of AMTA Workshop on MT Research and the
  Translation Industry}, Boston, US.

\bibitem[{Vaswani et~al.(2017)Vaswani, Shazeer, Parmar, Uszkoreit, Jones,
  Gomez, Kaiser, and Polosukhin}]{transformer}
Ashish Vaswani, Noam Shazeer, Niki Parmar, Jakob Uszkoreit, Llion Jones,
  Aidan~N Gomez, {\L}ukasz Kaiser, and Illia Polosukhin. 2017.
\newblock Attention is all you need.
\newblock In \emph{Advances in Neural Information Processing Systems 30}, pages
  6000--6010. Curran Associates, Inc.

\bibitem[{Wang et~al.(2017)Wang, Lu, Tu, Li, Xiong, and
  Zhang}]{hybrid-smt-advise}
Xing Wang, Zhengdong Lu, Zhaopeng Tu, Hang Li, Deyi Xiong, and Min Zhang. 2017.
\newblock Neural machine translation advised by statistical machine
  translation.
\newblock In \emph{AAAI}, pages 3330--3336.

\bibitem[{Wu et~al.(2016)Wu, Schuster, Chen, Le, Norouzi, Macherey, Krikun,
  Cao, Gao, Macherey et~al.}]{gnmt}
Yonghui Wu, Mike Schuster, Zhifeng Chen, Quoc~V Le, Mohammad Norouzi, Wolfgang
  Macherey, Maxim Krikun, Yuan Cao, Qin Gao, Klaus Macherey, et~al. 2016.
\newblock Google's neural machine translation system: Bridging the gap between
  human and machine translation.
\newblock \emph{arXiv preprint arXiv:1609.08144}.

\bibitem[{Zhou et~al.(2017)Zhou, Hu, Zhang, and Zong}]{hybrid-nmt-multisource}
Long Zhou, Wenpeng Hu, Jiajun Zhang, and Chengqing Zong. 2017.
\newblock Neural system combination for machine translation.
\newblock In \emph{Proceedings of the 55th Annual Meeting of the Association
  for Computational Linguistics (Volume 2: Short Papers)}, pages 378--384.
  Association for Computational Linguistics.

\end{thebibliography}
\bibliographystyle{acl_natbib_nourl}

\appendix

\end{document}